\documentclass[runningheads]{llncs}

\newcommand {\ent} {\mathrel{{\scriptstyle\mid\!\sim}}}
\newcommand {\imp} {\rightarrow}

\newcommand {\emme} {\mathcal{M}}
\newcommand {\enne} {\mathcal{N}}

\newcommand {\tc} {\mid}
\newcommand {\vuoto} {\emptyset}

\newcommand{\tip}{{\bf T}}

\newcommand{\lc}{\mathcal{LC}}
\newcommand{\alc}{\mathcal{ALC}}

\newcommand{\alct}{\mathcal{ALC}+\tip}

\newcommand{\el}{\mathcal{EL}}
\newcommand{\elpb}{{\mathcal{EL}}^{+}_{\bot}}

\newcommand{\be}{\begin{enumerate}}
\newcommand{\ee}{\end{enumerate}}

\newcommand{\hide}[1]{}

\def \cases{\left \{\begin{array}{l}}
\def \endcases{\end{array}\right .}

\newcommand {\ri} {\rightarrow}

\newcommand {\bes} {\begin{description}}
\newcommand{\ens} {\end{description}}
\newcommand {\la} {\langle}
\newcommand {\ra} {\rangle}

\newcommand {\beq} {\begin{quote}}
\newcommand {\enq} {\end{quote}}
\newcommand {\bit} {\begin{itemize}}
\newcommand {\enit} {\end{itemize}}

\newenvironment{pozz}{\color{black}}{\color{black}}
\usepackage{times}
\usepackage{helvet}
\usepackage{courier}
\usepackage{times}
\usepackage{undertilde}
\usepackage{graphicx}
\usepackage{latexsym}
\usepackage{graphicx}
\usepackage{color}
\usepackage{amssymb}
\usepackage{colortbl}
\usepackage{amsmath}
\usepackage{microtype}
\usepackage{amsmath}
\usepackage{tikz}

\def \ri{\rightarrow}

\begin{document}
\bibliographystyle{plain}

\title{From Common Sense Reasoning to \\ Neural Network Models through Multiple Preferences:\\
an overview}



\author{Laura Giordano \inst{1} \and Valentina Gliozzi \inst{2} \and Daniele Theseider Dupr{\'{e}}  \inst{1}}

\institute{DISIT - Universit\`a del Piemonte Orientale, 
 Alessandria, Italy 
 \and
Center for Logic, Language and Cognition \& 
Dipartimento di Informatica, \\
Universit\`a di Torino, Italy, 
}

\authorrunning{ }
\titlerunning{ }

 \maketitle
 

\begin{abstract}
In this paper we discuss the relationships between conditional and preferential logics and neural network models, based on a multi-preferential semantics. 
We propose a concept-wise multipreference semantics, recently introduced for defeasible description logics to take into account preferences with respect to different concepts, 
as a tool for  providing a semantic interpretation to neural network models.  
This approach has been explored both for unsupervised neural network models (Self-Organising Maps) and for supervised ones (Multilayer Perceptrons), and we expect that 
the same approach might be extended to other neural network models.
It allows for logical properties of the network to be checked (by model checking) over an interpretation capturing the input-output behavior of the network.
For Multilayer Perceptrons, the deep network itself can be regarded as a conditional knowledge base, 
in which synaptic connections correspond to weighted conditionals.
The paper describes the general approach, through the cases of Self-Organising Maps and Multilayer Perceptrons, and discusses some open issues and perspectives.
\end{abstract}

\section{Introduction}

Preferential approaches  \cite{KrausLehmannMagidor:90,Pearl90,whatdoes} to common sense reasoning, having their roots in conditional logics \cite{Lewis:73,Nute80},
have been recently extended to description logics, to deal with inheritance with exceptions in ontologies,
allowing for non-strict forms of inclusions,
called {\em typicality or defeasible inclusions} (namely, conditionals), with different preferential semantics \cite{lpar2007,sudafricaniKR} 
and closure constructions \cite{casinistraccia2010,CasiniDL2013,AIJ15,Pensel18}, 
allowing for defeasible or typicality inclusions, e.g., of the form $\tip(C) \sqsubseteq D$, meaning ``the typical $C$s are $D$s" or ``normally $C$s are $D$s",
corresponding, in the propositional case, to the conditionals $C \ent D$ in Kraus, Lehmann and Magidor's (KLM) preferential approach \cite{KrausLehmannMagidor:90,whatdoes}. Description logics allow for a limited first-order language. A first-order extension of system Z has also been explored \cite{BeierleJAR17}.

In this paper we consider  ``concept-wise" a multi-preferential semantics, recently introduced by Giordano and Theseider Dupr{\'{e}} \cite{TPLP2020} to capture preferences with respect to different aspects (concepts) in ranked $\el$ knowledge bases,
and describe how it has been used as a semantics for some neural network models. 
We have considered both an unsupervised model, Self-Organising Maps, and a supervised one, Multilayer Perceptrons. 

Self-organising maps (SOMs) are psychologically and biologically plausible neural network models \cite{kohonen2001}  that can learn after limited exposure to positive category examples, without need of contrastive information. They have been proposed as possible candidates to explain the psychological mechanisms underlying category generalisation. Multilayer Perceptrons (MLPs) \cite{Haykin99} are deep networks. Learning algorithms in the two cases  are quite different but, in this work, we only aim to capture, through a semantic interpretation, the behavior of the network resulting after training and not to model learning. We will see that this can be accomplished in both cases in a similar way, based on a multi-preferential semantics.

The result of the training phase is represented very differently in the two models: for SOMs it is given by a set of units spatially organized in a grid (where each unit $u$ in the map is associated with a
weight vector $w_u$ of the same dimensionality as the input vectors); 
for MLPs, as a result of training, the weights of the synaptic connections have been learned. 
In both cases, considering the domain of all input stimuli presented to the network during training (or in the generalization phase),  one can build a semantic interpretation describing the input-output behavior of the network as a multi-preference interpretation, where preferences are associated to concepts. For SOMs, the learned categories are regarded as concepts $C_1,\ldots, C_n$ so that a preference relation (over the domain of input stimuli) is associated to each category. In case of MLPs, each neuron in the deep network (including hidden neurons) is associated to a concept and a preference relation is associated to it.


In both cases, the  preferential model resulting from the network after training  
describes the input-output behavior of the network 
on the input stimuli considered, and the preference relations define a notion of typicality (with respect to different concepts/categories) on the domain of input stimuli. 
For instance, given two input stimuli $x$ and $y$, the model can assign to $x$ a degree of typicality which is higher than the degree of typicality of $y$ with respect to some category $\mathit{Horse}$, so that $x$ is regarded as a being more typical than $y$ as a horse 
($\mathit{x <_{Horse} y}$), while vice-versa $y$ can be regarded as a being more typical than $x$ as a zebra 
($\mathit{y <_{Zebra} x}$).
The preferential interpretation can be used for checking properties like: are the instances of a category $C_1$ also instances of category $C_2$? are  typical instances of a category $C_1$ also instances of category $C_2$? 
This verification can be done by {\em model-checking} given multipreference interpretation describing the input-output behavior of the network \cite{arXivSI_JLC2021}. 

This kind of construction establishes a strong relationship between the logics of commonsense reasoning and the neural network models, as the first ones are able to reason about the properties of the second ones. The relationship can be made even stronger in some cases, e.g., for MLPs, when the neural network itself can be seen as a 
conditional knowledge base.
In  \cite{JELIA2021}, the concept-wise multipreference semantics has been adapted to deal with weighted knowledge bases, where typicality inclusions  have a weight, a real (positive or negative) number, representing the plausibility of the typicality inclusions. 
It has been proven that Multilayer Perceptrons can be regarded  as weighted conditional knowledge bases under a fuzzy extension of the multipreference semantics. The multipreference interpretation which can be built over the  set of input stimuli to describe the input-output behavior of the deep network can be proven to be a coherent fuzzy multipreference model of such a knowledge base (under some condition on the activation functions).

This approach rises several issues, from the standpoint of knowledge representation,  from the standpoint of neuro-symbolic integration, as well as from the standpoint of 
explainable AI \cite{Adadi18,Guidotti2019,Arrieta2020}. We will discuss some of these issues in the paper after describing the approach in some detail.

\section{A concept-wise multi-preference semantics} \label{sec:multipref}

In this section  we shortly describe an extension of $\alc$ with typicality based on the same language as the typicality logics  \cite{lpar2007,AIJ15}, but
on a different concept-wise multipreference semantics first introduced for $\elpb$ \cite{TPLP2020}.

We consider the description logic $\alc$ 
Let ${N_C}$ be a set of concept names, ${N_R}$ a set of role names
  and ${N_I}$ a set of individual names.  
The set  of $\alc$ \emph{concepts} can be
defined as follows: 
  $C \ \ := A \tc \top \tc \bot \tc \neg C  \tc C \sqcap C \tc C \sqcup C \tc \exists r.C  \tc \forall r.C $, 
where $a \in N_I$, $A \in N_C$ and $r \in N_R$. 
A knowledge base (KB) $K$ is a pair $({\cal T}, {\cal A})$, where ${\cal T}$ is a TBox and
${\cal A}$ is an ABox.
The TBox ${\cal T}$ is  a set of {\em concept inclusions} (or subsumptions) of the form $C \sqsubseteq D$, where $C,D$ are concepts.
The  ABox ${\cal A}$ is  a set of assertions of the form $C(a)$ 
and $r(a,b)$ where $C$ is a  concept, $r \in N_R$, and $a, b \in N_I$.

In addition to standard $\alc$ inclusions $C \sqsubseteq D$ (called  {\em strict} inclusions in the following), the TBox ${\cal T}$  also contains typicality inclusions of the form $\tip(C) \sqsubseteq D$, where $C$ and $D$ are $\alc$ concepts and $\tip$ is a new concept constructor (and $\tip(C)$ is called a typicality concept).
A typicality inclusion $\tip(C) \sqsubseteq D$ means that ``typical $C$s are $D$s" or ``normally $C$s are $D$s" and corresponds to a conditional implication $C \ent D$ in Kraus, Lehmann and Magidor's (KLM) preferential approach \cite{KrausLehmannMagidor:90,whatdoes}. 
Such inclusions are defeasible, i.e.,  admit exceptions, while 
strict inclusions must be satisfied by all domain elements.

Let ${\cal C}= \{C_1, \ldots, C_k\}$ be a set of distinguished $\el$ concepts. 
For each concept $C_i \in {\cal C}$, we introduce a modular preference relation $<_{C_i}$ which describes the preference among domain elements with respect to $C_i$.
Each preference relation $<_{C_i}$ has the same properties of preference relations in KLM-style ranked interpretations \cite{whatdoes}, i.e., it is a modular and well-founded strict partial order (an irreflexive and transitive relation), where: $<_{C_i}$ is {\em well-founded} 
if, for all $S \subseteq \Delta$, if $S\neq \emptyset$, then $min_{<_{C_i}}(S)\neq \emptyset$;
and  $<_{C_i}$ is {\em modular} if,
for all $x,y,z \in \Delta$, if $x <_{C_j} y$ then ($x <_{C_j} z$ or $z <_{C_j} y$).

  \begin{definition}[Multipreference interpretation]\label{defi:multipreference}  
A {\em multipreference interpretation}  is a tuple
${\emme}= \langle \Delta, <_{C_1}, \ldots, <_{C_k}, \cdot^I \rangle$, 
where:\ \ $(a)$ \ $\Delta$ is a non-empty domain;

\begin{itemize}
 
\item[(b)] $<_{C_i}$ is an irreflexive, transitive, well-founded and modular relation over $\Delta$;

\item[(d)]  
$\cdot^I$ is an interpretation function, as in an $\el$ interpretation  
 that maps each
concept name $C\in N_C$ to a set $C^I \subseteq  \Delta$, each role name $r \in N_R$
to  a binary relation $r^I \subseteq  \Delta \times  \Delta$,
and each individual name $a\in N_I$ to an element $a^I \in  \Delta$.
It is extended to complex concepts  as follows:
$\top^I=\Delta$, $\bot^I=\vuoto$, 
$(\neg C)^I=\Delta \backslash C^I$, 
$(C \sqcap D)^I =C^I \cap D^I$  and $(C \sqcup D)^I=C^I \cup D^I$, 
$(\exists r.C)^I =\{x \in \Delta \tc \exists y.(x,y) \in r^I \ \mbox{and} \ y \in C^I\}$ and
$(\forall r.C)^I =\{x \in \Delta \tc \forall y. (x,y) \in r^I \imp y \in C^I\} $.
\end{itemize}

\end{definition}
The preference relation $<_{C_i}$ allows the set of prototypical  $C_i$-elements to be defined as the $C_i$-elements which are minimal with respect to $<_{C_i}$, i.e., $min_{<_{C_i}} (C_i^I)$.
As a consequence, the multipreference interpretation above is able to single out the typical $C_i$-elements, for all distinguished concepts $C_i \in {\cal C}$.

The multipreference  structures above are at the basis of the semantics for ranked $\el$ knowledge bases  \cite{TPLP2020}, which have been
inspired by Brewka's framework of basic preference descriptions  \cite{Brewka04}. 
While we refer to  \cite{TPLP2020} for the construction of the preference relations $<_{C_i}$'s 
from a ranked knowledge base $K$, in the following we shortly recall the notion of concept-wise multi-preference interpretation which can be obtained 
by {\em combining} the preference relations $<_{C_i}$ into a global preference relation $<$. 
This is needed for reasoning about the typicality of arbitrary $\el$ concepts $C$, 
which do not belong to the set of distinguished concepts ${\cal C}$.
For instance,  we may want to verify whether typical employed students are young, or whether they have a boss, starting from a ranked KB containing  inclusions $\mathit{\tip(Stud) \sqsubseteq Young}$, $\mathit{\tip(Emp) \sqsubseteq Has\_Boss}$, $\mathit{\tip(Emp) \sqsubseteq NonYoung}$, and
$\mathit{Young \sqcap NonYoung \sqsubseteq \bot}$. 
To answer the query above both preference relations $<_{\mathit{Emp}}$ and $<_{\mathit{Stud}}$ are relevant, and they might be conflicting 
as, for instance, Tom is more typical than Bob as a student ($\mathit{tom <_{\mathit{Stud}} bob}$), but more exceptional as an employee ( $\mathit{bob <_{\mathit{Emp}} tom}$).
By {\em combining} the preference relations $<_{C_i}$ into a single {\em global preference} relation $<$
we can exploit $<$ for interpreting the typicality operator, which may be applied to arbitrary concepts, and verify, for instance, whether 
$\mathit{\tip(Stud \sqcap Emp) \sqsubseteq Has\_Boss}$.

A natural definition of the notion of global preference $<$ exploits Pareto combination of the relations $<_{C_1}, \ldots,<_{C_k}$,
as follows:
\begin{align*}
x <y  \mbox{ iff \ \ } 
(i) &\  x <_{C_i} y, \mbox{ for some } C_i \in {\cal C}, \mbox{ and } \\
(ii) & \ \mbox{  for all } C_j\in {\cal C}, \;  x \leq_{C_j} y  
\end{align*}
where $\leq_{C_i}$ is the non-strict preference relation associated with $<_{C_i}$ ($\leq_{C_i}$ is a total preorder).
A slightly more sophisticated notion of preference combination, which exploits a modified  Pareto condition
taking into account the specificity relation among concepts (such as, for instance, the fact that concept $\mathit{PhdStudent}$ is more specific than concept $\mathit{Student}$), has been considered for ranked knowledge bases \cite{TPLP2020}. 

The addition of the global preference relation allows for defining a notion of {\em concept-wise multipreference interpretation} $\emme= \langle \Delta, <_{C_1}, \ldots,<_{C_k}, <, \cdot^I \rangle$, where typicality concept $\tip(C)$ is interpreted as the set of the $<$-minimal  $C$ elements,  i.e., $(\tip(C))^I = min_{<}(C^I)$,
where $Min_<(S)= \{u: u \in S$ and $\nexists z \in S$ s.t. $z < u \}$.

The notions of cw$^m$-model of a ranked $\el$ knowledge base $K$, and of cw$^m$-entailment can be defined in the natural way.
In particular, cw$^m$-entailment has been proven to be $\Pi^p_2$-complete for  $\elpb$ ranked knowledge bases and
 to satisfy the KLM postulates of a preferential consequence relation \cite{TPLP2020}.



\section{A multi-preferential interpretation of Self-organising maps}\label{sec:som-all}

In this section, we report about the multi-preferential semantics for SOMs proposed 
in \cite{CILC2020}, and later extended to fuzzy interpretations and to probabilistic interpretations in \cite{arXivSI_JLC2021}.


%
%

Self-organising maps, introduced by Kohonen \cite{kohonen2001}, are particularly plausible neural network models that learn in a human-like manner. 
%
In this section we shortly describe the architecture of SOMs and report about Gliozzi and Plunkett's  similarity-based account of category generalization based on SOMs \cite{CogSci2017}. 

SOMs  consist of a set of neurons, or units, spatially organized in a grid \cite{kohonen2001}. 
%
%
Each map unit $u$ is associated with a world representation, given by a weight vector $w_u$ of the same dimensionality as the input vectors.  
At the beginning of training, all weight vectors are initialized to random values, outside the range of values of the input stimuli. 
During training, the input elements  are sequentially presented to all neurons of the map. After each presentation of an input $x$, the {\em best-matching unit} (BMU$_x$) is selected: this is the unit $i$ whose weight vector $w_i$ is closest to the stimulus $x$ (i.e. $i = \arg\min_j\|x - w_j\|$).

The weights of the best matching unit and of its surrounding units are updated in order to maximize the chances that the same unit (or its surrounding units) will be selected as the best matching unit for the same stimulus or for similar stimuli {on subsequent presentations}. In particular,  it reduces the distance between the best matching unit's weights (and its surrounding neurons' weights) and the incoming input. 
The learning process is incremental: after the presentation of each input, the map's representation of the input  (in particular the representation of its best-matching unit) is updated in order to take into account the new incoming stimulus.
%
At the end of the whole process, the SOM has learned to organize the stimuli in a topologically significant way: similar inputs (with respect to Euclidean distance) are mapped to close by areas in the map, whereas inputs which are far apart from each other are mapped to distant areas of the map.

Once the SOM has learned to categorize, to assess category generalization, Gliozzi and Plunkett \cite{CogSci2017} define the map's disposition to consider a new stimulus $y$ as a member of a known category $C$ as a function of the {\em distance} of $y$ from the {\em map's representation} of $C$.
They use $BMU_{C}$ to refer to the map's representation of category $C$ and define category generalization as depending on 
 the distance of the new stimulus $y$ with respect to the category representation 
{\em compared to} the maximal distance from that representation of all known instances of the category.
This is captured by the following notion of {\em relative distance} ({\em rd} for short)  \cite{CogSci2017} :
\begin{equation}
\label{relative-distance}
rd(y,C) = \frac{min \|y - BMU_{C}\| }{max_{x \in C} \| x - BMU_{x}\| }
\end{equation}
where $min \|y - BMU_{C}\|$ is the (minimal) Euclidean distance between $y$ and $C$'s category representation, and ${max_{x \in C} \| x - BMU_{x}\| }$ expresses the {\em precision} of category representation,
and is the (maximal) Euclidean distance between any known member of the category and the category representation.

By judging a new stimulus as belonging  to a category by comparing the distance of the stimulus from the category representation to the precision of the category representation, Gliozzi and Plunkett demonstrate  \cite{CogSci2017}  that the Numerosity and Variability effects of category generalization, described by Griffiths and Tenenbaum \cite{tengrif2001}, and usually explained with Bayesian tools, can be {accommodated} within a simple and psychologically plausible similarity-based account. 
Their notion of relative distance can as well be used as a basis for a logical semantics for SOMs.  

%


\subsection{Relating self-organising Maps and multi-preference models}

Once the SOM has learned to categorize, we can regard the result of the categorization as a multipreference interpretation.  
Let $X$ be the set of input stimuli from different categories, $C_1, \ldots, C_k$, which have been considered during the learning process.
For each category $C_i$, we let $BMU_{C_i}$ be the ensemble of best-matching units corresponding to the input stimuli of category $C_i$, i.e.,
$BMU_{C_i}= \{ BMU_x \mid x \in X \mbox{ and } x \in C_i\}$.
We regard the learned categories $C_1, \ldots, C_k$ as being the concept names (atomic concepts) in the description logic and we let them constitute our set of distinguished concepts ${\cal C}= \{C_1, \ldots, C_k\}$. 

To construct a multi-preference interpretation, 
first we fix the {\em domain} $\Delta^{s}$ to be the space of all possible stimuli; 
then,  for each category (concept) $C_i$,  we define a preference relation $<_{C_i}$, exploiting the notion of relative distance of a stimulus $y$ from the map's representation of $C_i$. Finally, we define the interpretation of concepts.

%
%
%
%
Let $\Delta^{s}$ be the set of all the possible stimuli, including all input stimuli ($X \subseteq \Delta^s$) as well as the best matching units of input stimuli (i.e., $\{BMU_x \mid x \in X \} \subseteq \Delta^s$).  For simplicity, we will assume the space of input stimuli to be finite.

Once the SOM has learned to categorize, the notion of relative distance $rd(x,C_i)$ of a stimulus $x$ from a category $C_i$ 
can be used to build 
a binary preference relation $<_{C_i}$ among the stimuli in $\Delta^s$ w.r.t. category $C_i$ as follows:  for all $x, x' \in \Delta^s$,
\begin{align}\label{preferenza_Ci}
x <_{C_i} x' \mbox{\ \  iff \ \ } rd(x,C_i) < rd(x' ,C_i)
\end{align}
Each preference relation $<_{C_i}$ is a strict partial order relation on $\Delta^s$.
The relation $<_{C_i}$ is also well-founded, as we have assumed $\Delta^{s}$ to be finite. 

We exploit this notion of preference 
to define a concept-wise multipreference interpretation associated with the SOM. 
We restrict the DL language to the fragment $\lc$ of $\alc$ (plus typicality), not admitting roles. 

 \begin{definition}[multipreference-model of a SOM]\label{modello_Som}  
The {\em multipreference-model of the SOM} is a multipreference interpretation 
$\emme^{s}= \langle \Delta^{s}, <_{C_1}, \ldots, <_{C_k}, \cdot^I \rangle$ 
such that:
\begin{itemize}
\item[(i)] $\Delta^{s}$ is the set of all the possible stimuli, as introduced above; 

\item[(ii)]
for each $C_i \in {\cal C}$, $<_{C_i}$ is the preference relation defined by equivalence (\ref{preferenza_Ci}).

\item[(iii)]  
 the interpretation function $\cdot^I$ is defined for concept names (i.e. categories) $C_i$ as: 
 $$C_i^I= \{y \in \Delta^s \mid rd(y,C_i) \leq rd_{max,C_i} \}$$
where $ rd_{max,C_i}$ 
is the maximal relative distance of an input stimulus $x \in C_i$ from category $C_i$, that is,
$rd_{max,C_i} = max_{x \in C_i} \{rd(x, C_i)\}$. 
The interpretation function $\cdot^I$ is extended to complex concepts 
in the fragment of $\lc$ according to Definition \ref{defi:multipreference}.

\end{itemize}
\end{definition}
Informally, we interpret  as $C_i$-elements those stimuli whose relative distance from category $C_i$ is not larger than the relative distance of any input exemplar belonging to category $C_i$.
Given $<_{C_i}$, we can identify the most typical $C_i$-elements  wrt $<_{C_i}$ 
as the $C_i$-elements whose relative distance from category $C_i$ is minimal, i.e., the elements in $min_{<_{C_i}}(C_i^I)$.
Observe that 
 the best matching unit $BMU_x$ of an input stimulus $x \in C_i$ is an element of $\Delta^s$.
As, for $y=BMU_x$,
$rd(y,C_i)$ is $0$, 
$BMU_{C_i} \subseteq min_{<_{C_i}}(C_i^I)$.

\subsection{Evaluation of concept inclusions by model checking} \label{sec:model_checking}

We have defined a multipreference interpretation $\emme^{s}$ where,  in the domain $\Delta^{s}$ of the possible stimuli, we are able to identify, for each category $C_i$,  the $C_i$-elements as well as the most typical $C_i$-elements wrt $<_{C_i}$.
We can exploit $\emme^s$ 
to verify which inclusions are satisfied by the SOM by {\em model checking}, i.e., by checking the satisfiability of inclusions over model $\emme^s$.  
This can be done both for strict concept inclusions of the form $C_i \sqsubseteq C_j$ and for defeasible inclusions of the form $\tip(C_i) \sqsubseteq C_j$, where $C_i$ and $C_j$ are concept names (i.e., categories), 
by exploiting a notion 
 of maximal relative distance of  $BMU_{C_i}$ from $C_j$, defined as
$rd( BMC_{C_i},C_j) = max_{x \in C_i} \{rd(BMU_x, C_j)\}$.

We refer to \cite{CILC2020,arXivSI_JLC2021} for details. Let us observe that checking the satisfiability of strict or defeasible inclusions on the SOM may be non trivial, depending on the number of input stimuli that have been considered in the learning phase, 
although from a logical point of view, this is just model checking.
Gliozzi and Plunkett have  considered self-organising maps that are able to learn from a limited number of input stimuli, although this is not generally true for all self-organising maps \cite{CogSci2017}. 

Note also that the multipreference interpretation $\emme^s$ introduced in Definition \ref{modello_Som} allows to determine the set of $C_i$-elements for all learned categories $C_i$
and to define the most typical $C_i$-elements, exploiting the preference relation $<_{C_i}$. 
Although, we are not able to define, for instance, the most typical $C_i \sqcap C_j$-elements just using single preferences,
starting from $\emme^s$, 
we can construct a concept-wise multipreference interpretation $\emme^{som}$ that combines the preferential relations in $\emme^{s}$ into a global preference relation $<$, and provides an intepretation to  all  typicality concepts as $\tip(C_i \sqcap C_j)$.
The interpretation $\emme^{som}$ can be constructed from $\emme^s$ according to the definition of the global preference in Section  \ref{sec:multipref}.

As an alternative to a multipreference semantics for  SOMs, a fuzzy semantics has also been considered \cite{arXivSI_JLC2021}, based on fuzzy Description Logics 
\cite{LukasiewiczStraccia09}, as well as a related probabilistic account exploiting  Zadeh's  probability of fuzzy events  \cite{Zadeh1968}.

Our work has focused on the multipreference  interpretation of a self-organising map after the learning phase. 
However, the state of the SOM during the learning phase can as well be represented as a multipreference model (in the same way).
During training, the current state of the SOM corresponds to a model representing the beliefs about the input stimuli considered so far (beliefs concerning the category of the stimuli).
One can  regard the category generalization process as a model building process and, in a way, as a belief change process.
%
For future work, it would be interesting to study the properties of  this notion of change and compare it with the notions of change studied in the literature  \cite{gardenfors,belief-revision,KatsunoMendelzon89,Katsuno-Sato:91}. 


\section{A multi-preferential interpretation of a deep neural network} \label{sec:multiulayer_perceptron}



Let us first recall from \cite{Haykin99} the model of a {\em neuron} as an information-processing unit in an (artificial) neural network. The basic elements are the following:
\begin{itemize}
\item 
a set of {\em synapses} or {\em connecting links}, each one characterized by a {\em weight}. 
We let $x_j$ be the signal at the input of synapse $j$ connected to neuron $k$, and $w_{kj}$ the related synaptic weight;
\item
the adder for summing the input signals to the neuron, weighted by the respective synapses weights: $\sum^n_{j=1} w_{kj} x_j$;
\item
an {\em activation function} for limiting the amplitude of the output of the neuron (typically, to the interval $[0,1]$ or $[-1,+1]$).
\end{itemize}
The sigmoid, threshold and hyperbolic-tangent functions are examples of activation functions.
A neuron $k$ can be described by the following pair of equations: $u_k= \sum^n_{j=1} w_{kj} x_j $, and $y_k=\varphi(u_k + b_k)$,
where  $x_1, \ldots, x_n$ are the input signals and $w_{k1}, \ldots,$ $ w_{kn} $ are the weights of neuron $k$; 
$b_k$ is the bias, $\varphi$ the activation function, and $y_k$ is the output signal of neuron $k$.
By adding a new synapse with input $x_0=+1$ and synaptic weight $w_{k0}=b_k$, one can write: 
$u_k= \sum^n_{j=0} w_{kj} x_j $, and  $y_k=\varphi(u_k)$,
where $u_k$ is called the {\em induced local field} of the neuron.
The neuron can be represented as a directed graph, where the input signals $x_1, \ldots, x_n$ and the output signal $y_k$ of neuron $k$ are nodes of the graph.
An edge from $x_j$ to $y_k$, labelled $w_{kj}$, means that  $x_j$ is an input signal of neuron $k$ with synaptic weight $w_{kj}$.  

Neural network models are classified by their synaptic connection topology. In a {\em feedforward} network the {architectural graph} is acyclic, while in a {\em recurrent} network it contains cycles. 
In a feedforward network neurons are organized in layers. In a {\em single-layer} network there is an input-layer of source nodes and an output-layer of computation nodes. In a {\em multilayer feedforward} network there is one or more hidden layer, whose computation nodes are called {\em hidden neurons} (or hidden units).
The source nodes in the input-layer supply the activation pattern ({\em input vector}) providing the input signals for the first layer computation units.
In turn, the output signals of first layer computation units provide the input signals for the second layer computation units, and so on, up to the final output layer of the network, which provides the overall response of the network to the activation pattern.
In a recurrent network at least one feedback
exists. 


\subsection{A (two-valued) multipreference interpretation of multilayer perceptrons}  \label{sec:sem_for_NN}

In the following, we do not put restrictions on the topology the network, and we consider a network ${\cal N}$ after training, when the synaptic weights $w_{kj}$ have been learned. 
We associate a concept name $C_i \in N_C$ to any unit 
$i$ in ${\cal N}$ (including input units and hidden units) 
and construct a multi-preference interpretation over a (finite) {\em domain $\Delta$} of input stimuli,
the input vectors considered so far, for training and generalization.  
In case the network is not feedforward, we assume that, for each input vector $v$ in $\Delta$, the network reaches a stationary state  \cite{Haykin99}, in which $y_k(v)$ is the activity level of unit $k$. 
In essence, we are not considering the transient behavior of the network, but rather  it behavior at stationary states. 

Let ${\cal C}= \{ C_1, \ldots, C_n\}$ be a subset of concepts in $N_C$, the concepts associated to the units we are focusing on (e.g., ${\cal C}$ might be associated to the set of output units, or to all units). 
We associate to ${\cal N}$ and $\Delta$ a (two-valued) concept-wise multipreference interpretation over the boolean fragment of $\alc$ (with no roles or individual names). 
\begin{definition}
The  {\em cw$^m$interpretation $\emme_{\enne}^\Delta=\langle \Delta,<_{C_1}, \ldots, <_{C_n}, <, \cdot^I \rangle$ over $\Delta$
for network ${\cal N}$}  wrt ${\cal C}$ is a cw$^m$-interpretation where: 
\begin{itemize}
\item
 the interpretation function $\cdot^I$ is defined for named concepts $C_k \in N_C$ as: $x \in C_k^I$ if  $y_k(x) \neq 0$, and $x \not \in C_k^I$ if  $y_k(x) = 0$;
 \item
  for $C_k \in {\cal C}$,  relation $<_{C_k}$ is defined for $x,x' \in \Delta$ as:
 $x <_{C_k} x'$ iff $y_k(x) > y_k(x')$, where $y_k(x)$ is the output signal of unit $k$ for input vectors $x$.
 \end{itemize}
\end{definition}
The relation $<_{C_k}$ is a strict partial order, and $\leq_{C_k}$ and $\sim_{C_k}$ are defined as usual. In particular,  $x \sim_{C_k} x'$ for $x,x' \not \in C_k^I$.
Clearly, the boundary between the domain elements which are in $C_k^I$ and those  which are not could be defined differently, 
e.g., by letting $x \in C_k^I$ if  $y_k(x) > 0.5$, and $x \not \in C_k^I$ if  $y_k(x) \leq 0.5$. This would require only a minor change in the definition of the $<_{C_k}$. 

This model provides a multipreference interpretation of the network $\enne$, 
based on the input stimuli considered in $\Delta$.
For instance, when the neural network is used for categorization and a single output neuron is associated to each category, each concept $C_h$ associated to an output unit $h$  corresponds to a learned category. If $C_h \in {\cal C}$, the preference relation $<_{C_h}$ determines the relative typicality of input stimuli wrt category $C_h$. This allows to verify typicality properties concerning categories,  such as $\tip(C_h) \sqsubseteq D$ (where $D$ is a boolean concept built from the named concepts in $N_C$), by {\em model checking} on the model $\emme_{\enne}^\Delta$. 

Evaluating properties involving hidden units might be of interest, although their meaning is usually unknown.  
In the well known Hinton's family example \cite{Hinton1986}, one may want to verify whether, normally, given an old Person 1 and relationship Husband, Person 2 would also be old, i.e., $\tip(Old_1 \sqcap Husband) \sqsubseteq Old_2$ is satisfied. 
Here, concept $Old_1$ (resp., $Old_2$) is  associated to a (known, in this case) hidden unit for Person 1 (and Person 2), while Husband is associated to an  input unit.

\subsection{From a two-valued to a fuzzy preferential interpretation of multilayer perceptrons}  \label{sec:fuzzy_sem_for_NN}

The definition of a fuzzy model of a neural network ${\cal N}$, under the same assumptions as in previous section 
is straightforward.
In a fuzzy DL interpretation $I=\langle \Delta, \cdot^I \rangle$ \cite{LukasiewiczStraccia09} concepts can be interpreted as fuzzy sets,
and the fuzzy interpretation function $\cdot^I$ assigns to each
concept  $C\in N_C$ a function  $C^I :  \Delta \ri [0,1]$.
For a domain element $x \in \Delta$, $C^I(x)$ represents the degree of membership of $x$ in concept $C$.

Let $N_C$ 
be the set containing a concept name $C_i$ for each unit $i$ in $\enne$, including hidden units.
Let us restrict to the boolean fragment of $\alc$ with no individual names.
A {\em fuzzy interpretation $I_{\enne}=\langle \Delta, \cdot^I \rangle$ for $\enne$}  \cite{JELIA2021} is defined as follows:
\begin{itemize}
\item[(i)]
$\Delta$ is a (finite) set of input stimuli;
\item[(ii)]
the interpretation function $\cdot^I$ is defined for named concepts $C_k \in N_C$ as: $C_k^I(x)= y_k(x)$, $ \forall x \in \Delta$;
where $y_k(x)$ is the output signal of neuron $k$, for input vector $x$.
\end{itemize}
The verification that a fuzzy axiom $\la C \sqsubseteq D \geq  \alpha \ra$ is satisfied in the model $I_{\enne}$, can be done based on satisfiability in fuzzy DLs, according to the choice of the t-norm and implication function. It requires $C_k^I(x)$ to be recorded for all $k=1,\ldots, n$ and $x \in \Delta$.
Of course, one could restrict $N_C$  to the concepts associated to input and output units in $\enne$, so to capture the input/output behavior of the network.

The fuzzy interpretation $I_{\enne}$ above, induces a preference relation over the domain $\Delta$ as, for all $x,x' \in \Delta$,
 $x <_{C_k} x'$ iff $y_k(x) > y_k(x')$.
 Based on this idea, 
 a fuzzy multipreference interpretation ${\emme^{f,\Delta}_{\enne}}=\langle \Delta,<_{C_1}, \ldots, <_{C_n}, \cdot^I \rangle$ over $\cal C$
can be associated to the network $\enne$ starting from $I_{\enne}$.
 In a fuzzy multipreference interpretation a typicality concept $\tip(C)$ can be interpreted as a crisp concept having the value $1$  for the minimal $C$-elements in the domain with respect to the preference relation $<_C$, and $0$ otherwise. This relation is well-founded if we restrict to finite models (as we do), or to witnessed models, as usual in fuzzy DLs \cite{LukasiewiczStraccia09}.
 
\section{Multilayer perceptrons as weighted conditional knowledge bases}

The three interpretations considered above for Multilayer Perceptrons describe the input-output behavior of the network, and allow for the verification of properties by model-checking. The last one, ${\emme^{f,\Delta}_{\enne}}$ is, in essence, a combination of the first two, and can be proved to be a model of the neural network $\enne$ when it is regarded as a weighted conditional knowledge base. 

In this section, we report the notion of a weighted conditional knowledge base for $\alc$ from \cite{JELIA2021}, and we describe how a weighted conditional knowledge base $K_{\enne}$ can be associated to a deep network ${\enne}$.
We give some hint about its two-valued and fuzzy multipreference semantics, and we refer to \cite{JELIA2021} for a detailed description.

%

\subsection{Weighted conditional knowledge bases}

Weighted $\alc$ knowledge bases are $\alc$ knowledge bases in which defeasible or typicality inclusions of the form $\tip(C) \sqsubseteq D$ are given a positive or negative weight (a real number).

A  {\em weighted $\alc$ knowledge base} $K$, over a set ${\cal C}= \{C_1, \ldots, C_k\}$ of distinguished $\alc$ concepts,
is a tuple $\langle  {\cal T}_{f}, {\cal T}_{C_1}, \ldots, {\cal T}_{C_k}, {\cal A}_f  \rangle$, where  ${\cal T}_{f}$  is a set of fuzzy $\alc$ inclusion axiom, ${\cal A}_f$ is a set of fuzzy $\alc$ assertions 
and
${\cal T}_{C_i}=\{(d^i_h,w^i_h)\}$ is a set of weighted typicality inclusions $d^i_h= \tip(C_i) \sqsubseteq D_{i,h}$, where each inclusion $d^i_h$ has a weight $w^i_h$, a real number.
The concepts $C_i$ occurring on the l.h.s. of some typicality inclusion $\tip(C_i) \sqsubseteq D$ are called {\em distinguished concepts}.
Arbitrary $\alc$ inclusions and assertions may belong to ${\cal T}_{f}$ and ${\cal A}_{f}$.

\begin{example} \label{exa:Penguin}
Consider the weighted knowledge base $K =\langle {\cal T}_{f},  {\cal T}_{Bird}, {\cal T}_{Penguin},$ 
$ {\cal A}_f \rangle$, over the set of distinguished concepts ${\cal C}=\{\mathit{Bird, Penguin}\}$, with empty ABox 
and with $ {\cal T}_{f}$ containing the inclusions $\mathit{Penguin \sqsubseteq  Bird} $ and $\mathit{Black \sqcap Grey  \sqsubseteq  \bot}$.
%
%
The weighted TBox ${\cal T}_{Bird} $ 
contains the following weighted defeasible inclusions: 

$(d_1)$ $\mathit{\tip(Bird) \sqsubseteq Fly}$, \ \  +20  \ \ \ \ \ \ \ \ \  \ \ \ \ \  

$(d_2)$ $\mathit{\tip(Bird) \sqsubseteq \exists has\_Wings. \top}$, \ \ +50

$(d_3)$ $\mathit{\tip(Bird) \sqsubseteq  \exists has\_Feather.\top}$, \ \ +50;

\noindent
${\cal T}_{Penguin}$ contains the defeasible inclusions:


$(d_4)$ $\mathit{\tip(Penguin) \sqsubseteq  Fly}$, \ \ - 70   \ \ \ \ \ \ \ \ \ \  

$(d_5)$ $\mathit{\tip(Penguin) \sqsubseteq Black}$, \ \  +50;

$(d_6)$ $\mathit{\tip(Penguin) \sqsubseteq Grey}$, \ \  +10;

%
%
%
%

\noindent
 The meaning is that a bird normally has wings, has feathers and flies, but having wings and feather (both with weight 50)  for a bird is more plausible than flying (weight 20), although flying is regarded as being plausible. For a penguin, flying is not plausible (inclusion $(d_4)$ has a negative weight -70), while  and being black or being grey are plausible properties of prototypical penguins, and $(d_5)$ and $(d_6)$ have both a positive weight, 50 and 10, respectively
(for a penguin being black is more plausible than being grey).
 
 \end{example}
A two-valued semantics for weighted DL knowledge bases 
has been defined by developing a semantic closure construction 
in the same spirit as Lehmann's lexicographic closure \cite{Lehmann95}, but 
more related to Kern-Isberner's semantics of c-representations \cite{Kern-Isberner01,Kern-Isberner2014}.
In c-representations, both  the sum of the weights of the verified conditionals  and the sum of the penalties of falsified conditionals are 
considered. Here, conditionals have a single (positive or negative) weight, but negative weights can be interpreted as penalties.
We consider a concept-wise construction, as we want to associate different (ranked) preferences to the different concepts. For an element $x$ in the domain $\Delta$,
and a concept $C_i$, the weight $W_i(x)$ of $x$ wrt $C_i$ is defined as the sum of the weights $w_h^i$ of the typicality inclusions $\tip(C_i) \sqsubseteq D_{i,h}$  in  ${\cal T}_{C_i}$ verified by $x$ (and is $-\infty$ when $x$ is not an instance of $C_i$).
From this notion of weight of an element wrt concept $C_i$, the  {\em preference relation $\leq_{C_i}$}  can be defined 
as follows:
for $x,y \in \Delta$, $x  \leq_{C_i}  y $ iff $W_i(x) \geq W_i(y)$.
The higher the weight of $x$ wrt $C_i$ the higher is its typicality relative to $C_i$.
This closure construction allows for the definition of {\em concept-wise multipreference interpretations}
as  in Section  \ref{sec:multipref}.

A similar construction has been adopted in the fuzzy case. Rather then summing weights $w_h^i$ of the typicality inclusions $\tip(C_i) \sqsubseteq D_{i,h} \in  {\cal T}_{C_i}$ verified in $I$, $W_i(x)$ is defined by summing the products $w_h^i \cdot D_{i,h}^I(x)$ for all $h$, thus considering the degree of membership of $x$ in each $D_{i,h}$ (a value in the interval $[0,1]$). 
Furthermore,
for fuzzy multipreference interpretations, a condition is needed to enforce the {\em coherence} of the values $C_i^I(x)$, defining the degree of membership of a domain element $x$ in a concept $C_i$ in a fuzzy interpretation $I$, with the weights $W_i(x)$, which are computed from the knowledge base (given $I$).
The requirement that,  for all $x,y \in \Delta$, $C_i^I(x) \geq C_i^I(y)$  iff  $W_i(x) \geq W_i(y)$ leads to the definition of {\em coherent fuzzy multipreference models} (cf$^m$-models) of the weighted conditional knowledge base. We refer to \cite{JELIA2021} for details.

\subsection{Mapping multilayer perceptrons to conditional knowledge bases}  \label{sec:NN&Conditionals} 


Let us now consider how a multilayer perceptron can be mapped to a weighted conditional knowledge base.
For each unit $k$, we consider all the units $j_1, \ldots, j_m$ whose output signals are the input signals 
of unit $k$, with synaptic weights $w_{k,{j_1}}, \ldots, w_{k,{j_m}}$.  Let $C_k$ be the concept name associated to unit $k$ and $C_{j_1}, \ldots, C_{j_m}$ the concept names associated to units $j_1, \ldots, j_m$, respectively.
For each unit $k$ the following set ${\cal T}_{C_k}$ of typicality inclusions is defined, with their associated weights:
\begin{quote}
$\tip(C_k) \sqsubseteq C_{j_1}$ with  $w_{k,{j_1}}$, \\
$\ldots$ ,\\
$\tip(C_k) \sqsubseteq C_{j_m}$ with  $w_{k,{j_m}}$.
\end{quote}
Given ${\cal C}$, the knowledge base  extracted from network ${\enne}$ is defined as the tuple: $K^{\enne} = \langle  {\cal T}_{strict},{\cal T}_{C_1}, \ldots,$ $ {\cal T}_{C_n}, {\cal A}  \rangle$, where $ {\cal T}_{strict}= {\cal A}=\emptyset$ and, for each $C_k \in {\cal C}$,  $K^{\enne} $ contains the set  ${\cal T}_{C_k}$ of weighted typicality inclusions associated to neuron $k$ (as defined above).
$K^{\enne}$ is a weighted knowledge base over the set  of distinguished concepts ${\cal C}= \{ C_1, \ldots, C_n\}$.
Given a network $\enne$, it can be proven that the interpretation
${\emme^{f,\Delta}_{\enne}}$ (see Section  \ref{sec:fuzzy_sem_for_NN}) is a cf$^m$-model of the  knowledge base $K^{\enne}$, provided the activation functions $\varphi$ of all units are monotonically increasing and have value in $(0,1]$.

\noindent
We refer to  \cite{arXiv_JELIA2020} for the proof. Under the given conditions on activation functions, that hold, for instance, for the sigmoid activation function,
for any choice of ${\cal C} \subseteq N_C$  and for any choice of the domain $\Delta$ of input stimuli (provided that they lead to a stationary state of $\enne$), the fm-interpretation ${\emme^{f,\Delta}_{\enne}}$  
is a coherent fuzzy multipreference model of the defeasible knowledge base $K^{\enne}$. 


This result can be further generalized by weakening the notion of coherence of a fuzzy multipreference interpretation to a notion of faithfulness considered in \cite{arXiv_ECSQARU2021} (called weak consistency in the technical report \cite{arXiv_JELIA2020}). It has been proven that, also in the fuzzy case, the concept-wise multipreference semantics has interesting properties and satisfies most of the KLM properties, depending of their reformulation 
and  on the on the fuzzy combination functions.

\section{Conclusions}

We have explored the relationships between a concept-wise multipreference semantics and two very different neural network models, Self-Organising Maps and Multilayer Perceptrons,
showing that a multi-preferential semantics can be used to provide a logical model of the network behavior after training. 
Such a model can be used to learn or to validate conditional knowledge from the empirical data used for training and generalization,
by model checking of logical properties. A two-valued KLM-style preferential interpretation with multiple preferences and a fuzzy semantics have been considered,
based on the idea of associating preference relations to categories (in the case of SOMs) or to neurons (for Multilayer Perceptrons). 
Due to the diversity of the two models we would expect that a similar approach might be extended to other neural network models and learning approaches.


Much work has been devoted, in recent years, to the combination 
 of neural networks and symbolic reasoning \cite{GarcezBG01,GarcezLG2009,GarcezGori2019}, leading to the definition of new computational models, such as Graph Neural Networks \cite{GarcezGori2020}, Logic Tensor Network \cite{SerafiniG16}, Recursive Reasoning Networks \cite{Lukasiewicz2020}, 
 neural-symbolic stream fusion \cite{PhuocEL21},
and to extensions of logic programming languages
with neural predicates \cite{DeepProbLog18,NeurASP2020}. 
Among the earliest  
systems combining logical reasoning and neural learning are the 
KBANN \cite{KBANN94} and the 
CLIP \cite{CLIP99} 
systems and Penalty Logic \cite{Pinkas95}, a non-monotonic reasoning formalism used to establish a correspondence with symmetric connectionist networks. The relationships between normal logic programs and connectionist network have been investigated by Garcez et al. \cite{CLIP99,GarcezBG01}
and by Hitzler et al. \cite{HitzlerJAL04}.

The correspondence between neural network models and fuzzy systems has been first investigated by Bart Kosko in his seminal work  \cite{Kosko92}.
In his view, ``at each instant the n-vector of neuronal outputs defines a fuzzy unit or a fit vector. Each fit value indicates the degree to which the neuron or element belongs to the n-dimentional fuzzy set." Our fuzzy interpretation of a multilayer perceptron regards, instead, each concept (representing a single neuron) as a fuzzy set. 
This is the usual way of viewing concepts in fuzzy DLs \cite{Straccia05,LukasiewiczStraccia08,BobilloStraccia16}, and we have used fuzzy concepts within a multipreference semantics based on a semantic closure construction, 
in the line of
Lehmann's semantics for lexicographic closure \cite{Lehmann95} and Kern-Isberner's c-representations \cite{Kern-Isberner01,Kern-Isberner2014}. 
The multipreference semantics we have introduced for weighted conditionals appears to be a relative of c-representations, which generate the world ranks as a sum of impacts of falsified conditionals,  \cite{Kern-Isberner01,Kern-IsbernerAMAI2004}.
We have further considered 
a semantics with multiple preferences, in order to make it concept-wise: each distinguished concept $C_i$ has its own set ${\cal T}_{C_i}$ of (weighted) typicality inclusions, and an associated preference relation $<_{C_i}$.
This allows a preference relation to be associated to each category (e.g., in the preferential interpretation of SOMs) or neuron (in a deep network).
Related semantics with multiple preferences have been proposed, starting from Brewka's framework of basic preference descriptions  \cite{Brewka04}, based on different approaches: 
in system ARS, as a refinement of System Z by Kern-Isberner and Ritterskamp \cite{IsbernerRitterskamp2010}, using techniques for handling preference fusion;
in $\alct$ (an extension of $\alc$ with typicality) by Gil \cite{fernandez-gil};
in a refinement of rational closure by Gliozzi \cite{GliozziAIIA2016}; 
by associating multiple  preferences to roles by Britz and Varzinczak \cite{Britz2018,Britz2019};
in ranked $\el$ knowledge bases  by Giordano and Theseider Dupr{\'{e}} \cite{TPLP2020};
in the first-order logic setting by Delgrande and Rantsaudis 
\cite{Delgrande2020};  and
in the MP-closure \cite{AIJ21}.

%

For Multilayer Perceptrons, the logical semantics is based on the representation of a deep neural network as a conditional knowledge base,
where conditional implications are associated to synaptic connections.
That a conditional logic, belonging to a family of logics which are normally used  
for hypothetical and counterfactual reasoning, for common sense reasoning, and for reasoning with exceptions, can be used for capturing reasoning in a deep neural network model is  rather surprising.  It suggests that slow thinking and fast thinking \cite{kahneman2011thinking} 
might be more related than expected.

Opening the black-box and recognizing that multilayer perceptrons can be seen as a set of conditionals, 
can be exploited as a possible basis for 
an integrated use of symbolic reasoning  and neural networks (at least for this neural network model). 
While a neural network, once trained, is able and fast in classifying the new stimuli (that is, it is able to do instance checking), all other reasoning services such as satisfiability, entailment and model-checking are missing. 
These capabilities would be needed for dealing with tasks combining empirical and symbolic knowledge, such as,  for instance:  to prove whether the network satisfies some (strict or conditional) properties; 
to learn the weights of a conditional knowledge base from empirical data and use it for inference; 
to combine defeasible inclusions extracted from a neural network with other defeasible or strict inclusions for inference.

To make these tasks possible, 
the development of proof methods for such logics is a preliminary step. 
%
In the two-valued case multipreference entailment is decidable for weighted $\el$ knowledge bases \cite{JELIA2021}, and proof methods for reasoning with weighted conditional knowledge bases in $\el$ could, for instance, exploit 
 Answer Set Programming (ASP) encodings of the  concept-wise multipreference semantics, 
an approach already considered \cite{TPLP2020} to achieve defeasible reasoning from ranked knowledge bases in {\em asprin} \cite{BrewkaAAAI15}.
In the fuzzy case, an open problem is whether the notion of fuzzy-multipreference entailment  is decidable   
(even for the small fragment of $\el$ without roles), and under which choice of fuzzy logic combination functions.
Undecidability results for fuzzy description logics with general inclusion axioms 
\cite{BaaderPenaloza11,CeramiStraccia2011,BorgwardtPenaloza12}   motivate the investigation of decidable approximations of fuzzy-multipreference entailment.

An interesting issue is whether the mapping of deep neural networks 
to weighted conditional knowledge bases can be extended to more complex neural network models, such as Graph neural networks \cite{GarcezGori2020}, 
or whether different logical formalisms and semantics would be needed.

Another issue is whether the fuzzy-preferential interpretation of neural networks can be related with 
the probabilistic interpretation of neural networks based on statistical AI. This is an interesting issue, as the fuzzy DL interpretations we have considered, where concepts are regarded as fuzzy sets, also suggest a probabilistic account 
based on  Zadeh's  probability of fuzzy events  \cite{Zadeh1968}. We refer to \cite{arXivSI_JLC2021} for some results concerning a probabilistic interpretation of SOMs and to \cite{arXiv_JELIA2020}  for a preliminary 
account for MLPs. 
A methodology for commonsense reasoning based on probabilistic conditional knowledge under the principle of maximum entropy (MaxEnt) has been developed by Kern-Isberner   \cite{Kern-Isberner98} starting from the propositional case. 
Wilhelm et al. \cite{WilhelmKFB19} have recently shown how to calculate MaxEnt distributions in a first-order setting by using typed model counting and condensed iterative scaling, and have explored the connection to Markov Logic Networks for drawing inferences.
A description logic with probabilistic conditionals $\alc^{ME}$ has also been proposed \cite{WilhelmIsberner19} based on this methodology.


\end{document}